\documentclass[conference]{IEEEtran}
\IEEEoverridecommandlockouts
\usepackage{cite}
\usepackage{amsmath,amssymb,amsfonts}
\usepackage{algorithmic}
\usepackage{graphicx}
\usepackage{textcomp}
\usepackage{xcolor}
\usepackage[ruled,vlined]{algorithm2e}
\usepackage{array,multirow,graphicx}

\include{pythonlisting}

\def\BibTeX{{\rm B\kern-.05em{\sc i\kern-.025em b}\kern-.08em
    T\kern-.1667em\lower.7ex\hbox{E}\kern-.125emX}}
\newcommand{\STAB}[1]{\begin{tabular}{@{}c@{}}#1\end{tabular}}

\begin{document}

\title{\emph{EnSyth}: A Pruning Approach to Synthesis of Deep Learning Ensembles\\
{\footnotesize \textsuperscript{}}
\thanks{\textsuperscript{\textcopyright} 2019 IEEE}
}

\author{\IEEEauthorblockN{Besher Alhalabi}
\IEEEauthorblockA{\textit{School of Computing} \\
\textit{Birmingham City University}\\
Birmingham, United Kingdom \\
besher.alhalabi@mail.bcu.ac.uk}
\and
\IEEEauthorblockN{Mohamed Medhat Gaber}
\IEEEauthorblockA{\textit{School of Computing} \\
\textit{Birmingham City University}\\
Birmingham, United Kingdom \\
mohamed.gaber@bcu.ac.uk}
\and
\IEEEauthorblockN{Shadi Basurra}
\IEEEauthorblockA{\textit{School of Computing} \\
\textit{Birmingham City University}\\
Birmingham, United Kingdom \\
shadi.basurra@bcu.ac.uk}
}

\maketitle

\begin{abstract}
Deep neural networks have achieved state-of-art performance in many domains including computer vision, natural language processing and self-driving cars. However, they are very computationally expensive and memory intensive which raises significant challenges when it comes to deploy or train them on strict latency applications or resource-limited environments. As a result, many attempts have been introduced to accelerate and compress deep learning models, however the majority were not able to maintain the same accuracy of the baseline models. In this paper, we describe \emph{EnSyth}, a deep learning ensemble approach to enhance the predictability of compact neural network's models. First, we generate a set of diverse compressed deep learning models using different hyperparameters for a pruning method, after that we utilise ensemble learning to synthesise the outputs of the compressed models to compose a new pool of classifiers. Finally, we apply backward elimination on the generated pool to explore the best performing combinations of models. On CIFAR-10, CIFAR-5 data-sets with LeNet-5, \emph{EnSyth} outperforms the predictability of the baseline model.
\end{abstract}


\section{Introduction} \label{intro}

Deep learning has gained considerable attention over the past decade because it has achieved remarkable success in various domains \cite{krizhevsky_imagenet_2012} \cite{collobert_natural_nodate} \cite{fridman_mit_2017}. It has become a dominant tool to learn and solve complex problems.   
The secret behind its success is mostly related to employing many hidden layers that learn different features hierarchically, with classification at the output layer typically performed by the soft-max function. Due to this large number of layers, deep neural networks are  generally complex and computationally expensive. For instance, VGG-16 model \cite{simonyan_very_2014} has around 138 millions of parameters and requires approximately 550 MB of memory space -- such a model could neither be trained nor deployed on low-memory devices. To overcome the complexity of deep neural networks, many compression and acceleration strategies have been introduced by the research community including (1) simple regularisers ($L1$ and $L2$) that could be used in the training process to control the complexity of a neural network \cite{nowlan_simplifying_1992} \cite{girosi_regularization_1995}; (2) DropConnect methods which produce pruned networks by randomly dropping a subset of weights \cite{wan_regularization_2013}, (3) neuron pruning methods that increase the sparsity of neural networks by removing irrelevant connections \cite{sietsma_neural_1988}\cite{hassibi_optimal_1993}\cite{hong-jie_xing_two-phase_2009}\cite{zhang_deep_2018}; (4) weight pruning that removes the weights that have less contribution to the network's output \cite{hagiwara_simple_1994}\cite{han_deep_2015}; (5) channel pruning which is another type of weight pruning but instead of removing a single neuron's output, it removes complete channels \cite{he_channel_2017}; (6) knowledge distillation that aims to transfer the knowledge from a teacher model (large network) into a student model (small network) \cite{romero_fitnets:_2014}\cite{you_learning_2017}\cite{yim_gift_2017}; (7) network quantisation that reduces the number of required bits to represent the network's weights, but it may require special hardware for acceleration\cite{gong_compressing_2014}\cite{wu_quantized_2016}; (8) Fine tuned network design, which falls into the acceleration category, that works toward reducing the complexity and improving the accuracy of the whole neural network. This is achieved  by the use of optimisation of the network's architecture and storage \cite{chollet_xception:_2016}\cite{howard_mobilenets:_2017}\cite{Kamada2018AdaptiveSL}; and (9) Genetic algorithms that have been applied widely in accelerating deep neural networks \cite{siebel_evolutionary_2007}\cite{xie_genetic_2017}\cite{hu_novel_2018}.
Only a few of the previous methods were able to produce compressed models with almost zero effect on the model's performance. 

In this paper, we propose \emph{EnSyth} an approach to synthesise deep learning ensembles from a baseline model, leading to boosting up its performance, in terms of accuracy and inference time. \emph{EnSyth} works by applying multiple sets of pruning methods by varying the corresponding hyperparameters. This, in turn, leads to a pool of diverse pruned deep learning models. Using such a pool to form deep learning ensembles, will have a powerful potential of boosting up the accuracy because of the awarded diversity from varying different values for pruning method's hyperparameters.
The number of possible ensembles that can be formed is $2^m-1$, where $m$ is the number of pruned models. Exploring such a large solution space for a large $m$ can be performed using a meth-heuristic optimisation method such as Genetic Algorithm \cite{kramer2017genetic}. In this paper, and to provide a proof of concept, we use a simple backward elimination method, where models of sizes $\{m, m-1, \dots, 1\}$ are formed by eliminating the worst performing pruning model sequentially. Accelerating inference is achieved through parallel processing, where all models can be used to infer the class label independently, and a computationally cheap fusion for majority voting is finally performed. These resultant ensembles can be particularly applicable in collaborative machine learning environments that operate on resource constrained devices \cite{gaber2013pocket}. 
The main contributions of this paper are as follows:
\begin{itemize}
    \item synthesis of an ensemble of pruned deep learning models from a baseline model from a diverse space of synthesised models. The diversity of the ensemble makes it possible to outperform a baseline model; and
    \item elimination of the impact of compression on deep learning models, by producing compressed models with better predictability measures. Through parallel processing, the approach achieved fast inference of the pruned models while boosting up the accuracy through ensembling.
\end{itemize}

The rest of this work is organised as follows; in section, \ref{related_work} we introduce related work in the field of accelerating and compressing of neural nets; in section, \ref{method} we move to illustrate in details the proposed technique \emph{EnSyth}. After that, we summarise the experiment on CIFAR-10, CIFAR-5, and MNIST-FASHION in section \ref{exp}; finally, we conclude this paper and present future works in section \ref{conclusion}.

\section{Related work} \label{related_work}
Recently, researchers have shown an increased interest in accelerating and compressing neural networks because of the emerging need for deploying those powerful models in resource-limited environments. A considerable amount of literature has been published and could be mainly divided into:
\subsection{Parameters pruning and sharing} this kind of techniques are widely used not only to reduce the complexity of a network but also to handle the over-fitting problems. \cite{lecun_optimal_1990} \cite{hassibi_optimal_1993} have introduced one of the early promising results on pruning neural networks where Hessian matrices had been applied to remove the redundant connections. \cite{han_learning_2015} proposed a method which aims to prune a network with nearly zero effect on the compressed model's performance. First, a training process is achieved to learn the important neurons that contribute much to the network; then the less important connections are removed. Finally, the network is retrained to fine-tune the weights of the network. The simulation on AlexNet \cite{krizhevsky_imagenet_2012} shows up to 9x compression ratio (from 61 million parameters to 6.7 million). However, most of those pruning techniques depend on the famous L1 and L2 regularisation \cite{nowlan_simplifying_1992} \cite{girosi_regularization_1995} which take a very long time to cover, besides, it needs a lot of efforts to calculate the sensitivity of the parameters. 

\subsection{Quantization} 
The main purpose of a quantisation process is to reduce the number of bits that are required to represent the weights of the neurons. \cite{gong_compressing_2014} and \cite{wu_quantized_2016} have used k-means scalar quantisation on neural network parameters (weights and bias) to reduce the complexity of the network. \cite{vanhoucke_improving_nodate} has applied 8-bits linear quantisation to speed up the training process without compromising on the accuracy. Deep compression \cite{han_deep_2015} has achieved the state of the art performance and reduced AlexNet model \cite{krizhevsky_imagenet_2012} by factor of 35× [240MB to 6.9MB]. Deep compression depends on \cite{han_learning_2015} to prune the less important connections then retrain the network. After that, it uses 8-bits quantisation to shorten the number of bits that are used to represent the weights. Finally, it applies Huffman coding to the quantised weights to gain more compression ratios. 
\subsection{Knowledge distillation}
\cite{bucilua_model_2006} has proved that it’s possible to transfer a compressed knowledge from an ensemble into a single model. \cite{hinton_distilling_2015} has extended the previous work to develop a new compression framework for neural nets. The proposed framework compresses a teacher model (ensemble of deep neural nets) into a student model by training the student to predict the output of the teacher. The proposed knowledge distillation work in \cite{romero_fitnets:_2014} aims to train shallow networks called FitNets from deep neural networks. FitNets allows to train deeper but thinner student model using the target outputs from the teacher, it distils the teacher model’s knowledge by minimising its' features map and pass it to the student. The experiments show the student has outperformed the teacher's performance. Although KD approaches could scientifically reduce the computation cost for deep neural nets, and may achieve a promising result in image classification tasks, it has limited use because it could be only applied on classification problems with SoftMax function. 
\subsection{Fine-tuned network design}
This category involves optimisation for the network's architecture and the convolutional layers' design (if applicable). \cite{howard_mobilenets:_2017} uses depth-wise convolutions to generate compressed models that could easily fit on mobile devices. Depth-wise convolutions were firstly introduced by \cite{sifre_rigid-motion_2014}, then it was adopted by \cite{ioffe_batch_2015} to present the Inception model. Inception model primarily based on co-variate shifts which is a useful technique in minimising the number of activation and reduce the training time. Similarly, \cite{chollet_xception:_2016} has introduced Xception. The theory behind this model could be illustrated as: 
First, the model tries to discover correlations between channels connections by a group of (1*1) convolution fitters. Later, it maps the input neurons to smaller input spaces (3 or 4). After that it maps any possible correlation in the smaller spaces by using $(3\times3)$/$(5\times5)$ convolutional filters.
\subsection{Evolutionary methods}
In \cite{d._whitley_evolution_1990}, the authors have utilised the genetic algorithms in pruning neural networks.Here, multiple versions of the pruned model are generated using reproduction, mutation and crossover. Similarly, Zhang \cite{zhang_evolving_1993} has developed an approach to simplify the architecture of a feed-forward neural network by applying genetic algorithms. The problem of finding the optimal architecture for a network is considered as a multi-objective optimisation, and the solution space ideally contains all possible combination of hidden nodes then GA is used to search for the perfect architecture.
\cite{hu_novel_2018} has presented a novel approach to prune CNN's using genetic algorithms, the method could be summarised as 1- a layer by layer pruning is proposed for a CNN model according to the sensitivity of each layer;2- tuning the pruned model using knowledge distillation framework;3- the channel selection process is formulated as a search problem which could be solved efficiently using genetics algorithms;4- two-step approximation fitness function is used to add extra efficiency to the genetic process. Generally, GA based pruning approaches are outperforming because they are improving the generalisation of the pruned network and for using fewer parameters.

\section{\emph{EnSyth}: Synthesis of Deep Learning Ensembles} \label{method}
In this section, we first introduce the topology of the feed-forward neural network models, then we explain the pruning method \cite{aghasi_fast_2018} that has been used to generate the compressed deep learning models. After that, we illustrate our approach to synthesis the compressed models, and the selection mechanism used to filter the best ones.
\subsection{Feed-forward neural networks}
Let's assume the training of a network is done using $x_p$ training example where:
\begin{enumerate}
    \item $p=1,\cdots,P$;
    \item $x_p \in \mathbb{R}^N$: $\mathbb{R}^N$ is the network's input;
\end{enumerate}
Suppose $X \in \mathbb{R}^{N \times P}$ is a one dimensional matrix represents the training samples as  $X = [x_1,\cdots,x_P]$, $L$ is a layer in the network; the network's output at the last layer is represented by $X^{(L)} \in \mathbb{R}^{N_L \times P}$ where each column in $X^{(L)}$ is a response to the corresponding training column in $X$.
In ReLU neural network, the output of $\ell^{th}$ layer is defined as :
\begin{equation}
 X^{(\ell)} = ReLU\Big(W_\ell^T X^{(\ell-1)}+ b^{(\ell)}1^T \Big)
\end{equation}
where $\ell=1,\cdots,L.$.

If we add an additional row to both of $X^{(\ell-1)}$ and $W_\ell$ the previous formula could be written as:
\begin{equation}
  X^{(\ell)} = ReLU \Big( W_\ell^TX^{(\ell-1)}\Big)
\end{equation}
where $\ell=1,\cdots,L.$.

A neural network which follows on of the two previous formulas(1,2) will be an ideal candidate for the pruning method which will be described next.

\subsection{Net-Trim- The Pruning method}

Net-Trim is a post-processing pruning framework, this means it prunes a network after a training process. It reduces the complexity of a given network by introducing more simple operations between the input/output of each layer while preserving the input data to be similar to the initial state (before pruning). The pruning process for each layer is considered as a convex optimisation problem that tries to convert large parts of the network's block to sparse matrices (most of the elements are zero). In simple words, to simplify the architecture of a neural network, Net-Trim removes the redundant connections and redirect the processing to a small group of important connections.


As seen before, in ReLU neural networks, the output of each layer could be defined as:
\begin{equation}
 X_{out} = ReLU \Big(W X_in\Big)
\end{equation}
where:
\begin{itemize}
    \item $X\_out$ is the output matrix;
    \item $W$ is the network weight matrix;
    \item $X\_in$ is the input matrix, each column in this matrix corresponds to a training sample;
\end{itemize}
Net-Trim simply replace $W$ with a spare matrix $U$ using convex variant of the following :
\begin{equation}
 \min_{\mathbf{U}}\|\mathbf{U}\|_1 ~ ~~~ \mbox{subject to:}~~~~ \mathbf{X}_{out} \approx \operatorname{ReLU}\Big(\mathbf{U}^\top\mathbf{X}_{in}\Big).   
\end{equation}
Net-Trim has different types of hyperparameters that have a direct effect on the generated models' accuracy and size. By choosing different values for those parameters, a diverse set of compressed models could be obtained. Each model in this set will have different characteristics from the others including accuracy, size and inference time. 
Our approach mainly depends on generating such a set that represents an optimal candidate for combining and synthesising the ensembles of deep learning models.
\subsection{Synthesis of deep learning ensembles}
Synthesising sets of divers compressed models into ensemble predictions is a critical element in our approach because ensemble learning will not only allow to create better classifiers but also overcome potential overfitting issues and provide more generalisation to the final solution. 
Let $m$ a pruned model generated by Net-Trim and the decision of the model $m_i$ about a class $w_j$ is defined as:
$y_{i,j} \in \{0,1\} $ where:
\begin{itemize}
    \item $i = 1,2,\cdots,N$ : $N$ is the number of the classifiers.
    \item $j = 1,2,\cdots,C$ : $C$ is the number of classes.
\end{itemize}
if $m_i$ predicted correctly a class $w_j$ then $y_{i,j} = 1$ otherwise $y_{i,j} = 0$.

We use plurality voting \cite{10.1007/3-540-45014-9_1} as a simple ensemble learning technique to synthesise the classifiers. The prediction of each compressed model is considered as a vote, then the predictions which get the majority of votes will be found in the final ensemble's prediction.

Let $P= {m_1,\cdots,m_N}$ an ensemble of pruned models $m_i$, the final ensemble decision about a test class in plurality voting is a class $w_j$ that receives the maximum support $\eta_{final} (P)$ from all the classifiers which form the ensemble, thus the output of the ensemble could be defined as:
$$\eta_{final} (P) = argmax_{j\in \{1,\cdots,C\}}\sum_{i=1}^N y_{i,j} $$

However, the ensemble will contain some models with low predictability levels which may have an adverse effect on the overall performance, so we suggest backward elimination scheme to remove the models with reduced predictability levels and find the optimal combinations that achieve better results.
\subsection{Backward elimination}
Our goal is to have accurate ensembles that have the lowest possible number of models without compromising the predictability of the ensembles. Preserving the number of models to the minimum in each ensemble will be a critical factor when it comes to deploying on resource-constrained devices. 
Backward elimination is an approximation process that begins with all variables included, then eliminates one variable at a time until a stopping condition is reached. 
In our case, the backward elimination process will start with a full set of pruned models, and in each round we compute the performance of overall ensemble then remove the model that has the lowest accuracy. The loop will stop when the ensemble size is equal to one. Algorithm.~\ref{algorithm1} describes backwards elimination procedure on a pool of deep learning models.

\begin{figure*}[htbp]
    \centering
    \includegraphics[width=\textwidth]{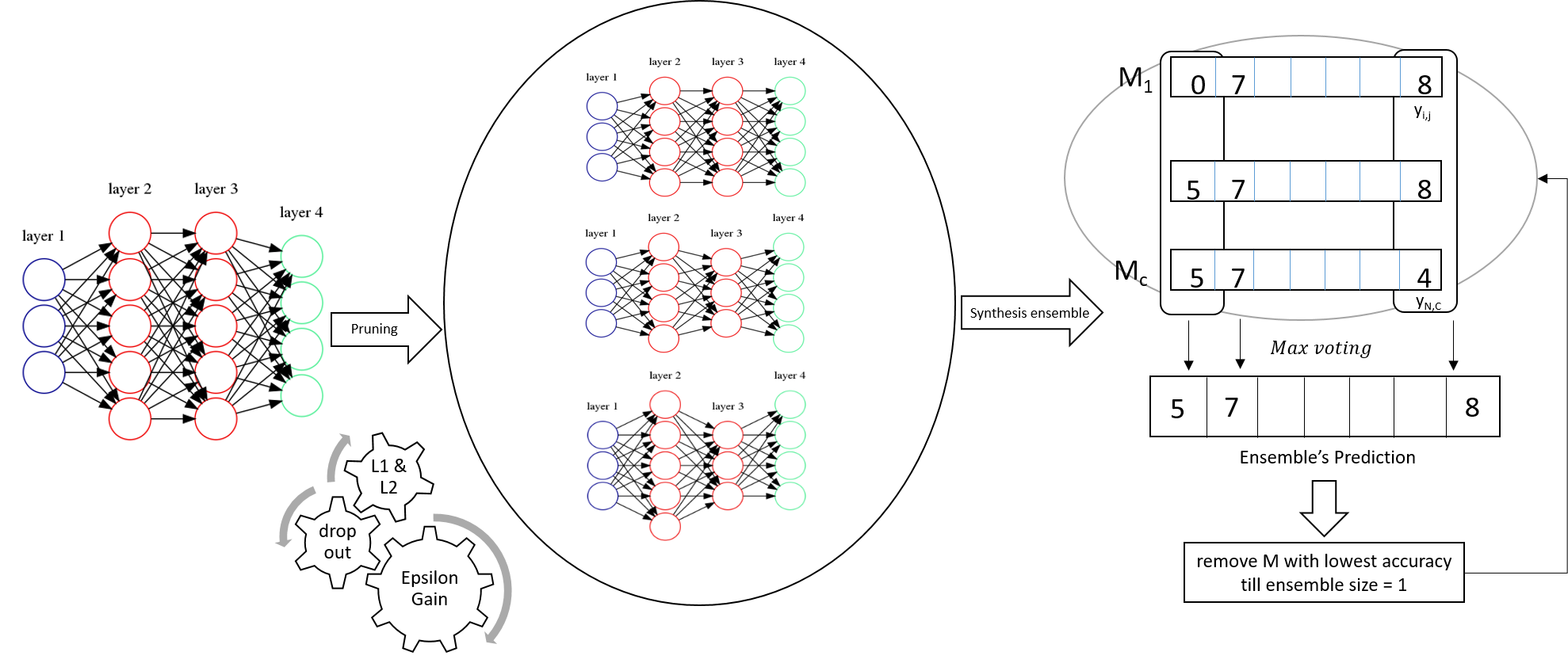}
    \caption{EnSyth, an approach to synthesis of deep learning ensembles}
    \label{method_fig}
\end{figure*}

\begin{algorithm}
\SetAlgoLined
 $N=Size(Ensemble)$\;
 \While{$N>0$}{
  $predict \gets \eta_{final} (P)$ \;
  Remove $m$ that has lowest accuracy from the $ensemble$;
  $N = N-1$ \;
 }
 \caption{Backward Elimination}
 \label{algorithm1}
\end{algorithm}
At the end of this process, we have a list of ensembles with the best possible combination of the models that achieved the highest accuracy.
Fig.~\ref{method_fig} illustrates \emph{EnSyth}, it could be summarised as:

\begin{itemize}
    \item train a baseline model;
    \item prune the baseline model with different values of the hyperparameters to formulate the solution's space;
    \item synthesis deep learning ensembles that belong to the existing space;
    \item apply backward elimination on the composed ensembles to select the best performing ensembles.
\end{itemize}

\section{Experiment} \label{exp}
In this section, the performance of our method is evaluated with LeNet-5 model on CIFAR10 and MNIST-FASHION datasets.
\subsection{LeNet-5}
LeNet-5 \cite{lecun_gradient-based_1998} is a popular convolutional neural network (CNN) model, it consists of two convolutional layers with (32*5*5) filter size for the first one and (64*5*5) filter size for the second one, both of them are followed by average pooling layers with (2*2) filter size and stride of two, and two fully connected layers. The architecture could be represented as: $INPUT => CONV => POOL => CONV => POOL => FC => FC$
where conv:Convolutional Layer, FC:Fully Connected layer, POOL:Average Pooling Layer

\subsection{Datasets}

\subsubsection{CIFAR-10}
CIAFR-10 \cite{krizhevsky_learning_nodate} is a benchmarking dataset with 70,000 images (28x28 colour) spanning over 10 categories of objects with 6,000 image per each category. CIFAR-10 has 50,000 images as a training set and 10,000 as a testing set. Each training and testing example is assigned to one of the following labels: [0: airplane, 1: automobile, 2: bird, 3: cat, 4: deer, 5: dog, 6: frog, 7: house, 8: ship, 9: truck]
\subsubsection{CIFAR-5}
We divided CIFAR-10 into a smaller subset for simulation purposes. CIFAR-5 consists of images spanning over five animal categories, which is 5 categories less than the original set. There are 6,000 images per each category, thus 25000 images were used for training and 5000 for testing. Each training and testing example is assigned  one of the following labels: [2: bird, 3: cat, 4: deer, 5: dog, 6: frog]
\subsubsection{MNIST-FASHION}
MNIST-FASHION \cite{xiao_fashion-mnist:_2017} is a new benchmarking dataset for machine learning problems and has 70,000 fashion products images spanning over ten categories with 7,000 images per each category. The training set has 60,000 images while the testing set has 10,000 images. All images are grayscale with a size of 28x28. Each training and testing example is assigned to one of the following labels: [0: T-shirt/top, 1: Trouser, 2: Pullover, 3: Dress, 4: Coat, 5: Sandal, 6: Shirt, 7: Sneaker, 8: Bag, 9: Ankle boot]
\subsection{Experimental setup}
Our experiment has been conducted on: OS: Ubuntu Desktop 18.04.1 LTS, CPU: Intel KVM 64bit 2,400 GHz, CASH: 16384 KB, RAM:32 GB, Python:3.6.7, TensorFlow:1.10.0, Keras:2.2.4.

\subsection{Network training and pruning}
We train LeNet-5 model on the proposed data sets to establish baseline models. The accuracy of LeNet-5 baseline model on CIFAR-10 is 78.4\% , CIFAR-5 is 73.3\% and 90.3\% on MNIST-FASHION. 
After that, we prune and fine tune the baseline models with Net-Trim \cite{aghasi_fast_2018}. Net-trim's has four hyperparameters: 
L1: apply L1 regularisation on model's weight;
L2: apply L2 regularisation on model's weight;
dropout: a factor used to ignore neurons during a training process randomly;
Epsilon\_gain: has a direct effect on the accuracy as well the sparsity of the pruned model.
We generate 36 compressed models providing different values for the previous parameters, Table.~\ref{tab1} shows in details the combinations of values that had been used to create a pool of compressed models.

\begin{table}[htbp]
\caption{hayperparameters values}
\begin{center}
\begin{tabular}{|c|c|c|c|c|}
\hline
\textbf{Solution}&\multicolumn{4}{|c|}{\textbf{Hyperparameters}} \\
\cline{2-5} 
\textbf{Space} & $\epsilon$& L1 & L2 & dropout\\
\hline
\multirow{12}{*}{\STAB{\rotatebox[origin=c]{90}{Set1}}}
 & 0.01 &  &  &  \\
 & 0.02 &  &  &  \\
 & 0.04 &  &  &  \\
 & 0.06 &  &  & \\
 & 0.08 &  &  &  \\
 & 0.1 & 0 & 1 & 1 \\
 & 0.2 &  &  &  \\
 & 0.3 &  &  &  \\
 & 0.4 &  &  &  \\
 & 0.5 &  &  &  \\
 & 0.6 &  &  &  \\
 & 0.7 &  &  &  \\
\hline
\multirow{12}{*}{\STAB{\rotatebox[origin=c]{90}{Set2}}}
 & 0.01 &  &  &  \\
 & 0.02 &  &  &  \\
 & 0.04 &  &  &  \\
 & 0.06 &  &  & \\
 & 0.08 &  &  &  \\
 & 0.1 & 0 & [0,0,0.004,0.004,0] & 1 \\
 & 0.2 &  &  &  \\
 & 0.3 &  &  &  \\
 & 0.4 &  &  &  \\
 & 0.5 &  &  &  \\
 & 0.6 &  &  &  \\
 & 0.7 &  &  &  \\
\hline
\multirow{12}{*}{\STAB{\rotatebox[origin=c]{90}{Set3}}}
 & 0.01 &  &  & \\
 & 0.02 &  &  &  \\
 & 0.04 &  &  &  \\
 & 0.06 &  &  & \\
 & 0.08 &  &  &  \\
 & 0.1 &  &  &  \\
 & 0.2 & 0 & [0,0,0.004,0.004,0.004]  &  0.5 \\
 & 0.3 &  &  &  \\
 & 0.4 &  &  &  \\
 & 0.5 &  &  &  \\
 & 0.6 &  &  &  \\
 & 0.7 &  &  &  \\
\hline
\end{tabular}
\label{tab1}
\end{center}
\end{table}
The above table represents the solution space for our method, but the real solution space may contain up to $2^{36}-1$ models where different combinations and permutations could be obtained. 
The following is a summary statistics about the size of the 36 pruned models where only the weights and biases are saved as compressed numpy arrays \cite{numpy_arrays}:
\begin{enumerate}
    \item CIFAR-10: Max:4.2 MB, Min:1.6 MB, Avg: 2.94 MB (baseline: 4.8 MB);
    \item CIFAR-5: Max: 4.2 MB, Min: 1.3 MB, Avg: 2.89 MB (baseline: 4.8 MB);
    \item MNIST-FASHION: Max:7.7 MB, Min: 0.88 MB, Avg: 2.56 MB (baseline: 7.7MB);
\end{enumerate}
\subsection{Synthesis of compressed deep learning ensembles}
The weights of models in the current solution space are combined to form new classifiers using majority voting. Next, the classifier's predictability is evaluated against the testing set. Then a simple backward elimination is applied to exclude the weakest model (i.e. the model with the lowest accuracy on the testing set). Ideally, the elimination process should be performed on a validation set, rather than the testing set. However, as the aim of this work is to prove the existence of pruned ensembles that are synthesised from the same baseline model, the testing set was used in the backward elimination process. The use of effective multi-objective optimisation methods over validation sets to find even more accurate and smaller ensembles is the next stage in this work, having our hypothesis about the existence of synthesised pruned ensembles proven true, as will be shown later in this section.     

\subsection{Results of CIFAR-10}
\begin{figure}[htbp]
    \centering
\includegraphics[width=\columnwidth]{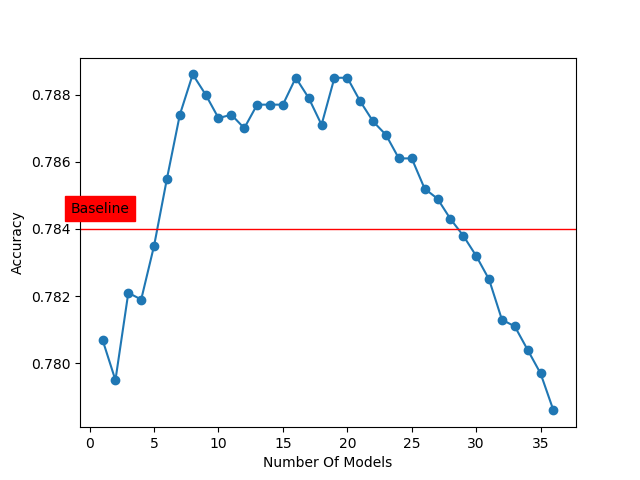}
    \caption{CIFAR10}
    \label{cifar10_results}
\end{figure}

Fig.~\ref{cifar10_results} summarises the results on CIFAR-10 testing set. As expected, our proposed method shows better predictability for the ensembles over the baseline model. Additionally, there is a clear trend of increased predictability where the number of models in a prediction ensemble is between [6--30]. A closer inspection to this figure reveals that 23  out of 36 models are outperforming the baseline model, and the highest  accuracy  achieved from an ensemble with 8 models is 78.86\%.

\subsection{Results of CIFAR-5}

\begin{figure}[htbp]
    \centering
\includegraphics[width=\columnwidth]{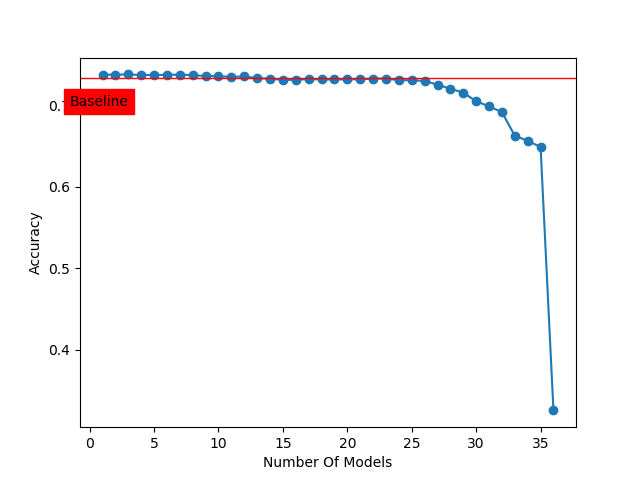}
    \caption{CIFAR5}
    \label{cifar5_results}
\end{figure}

From the data shown in Fig.~\ref{cifar5_results}, it is apparent that \emph{EnSyth} achieved a slightly better accuracy than the baseline model by synthesising deep learning ensembles. There are 12 outperforming ensembles found in the explored solution space (i.e. 36 solutions out of the possible $2^{36}-1$ solutions), and the highest accuracy obtained was 73.82\% as a result of combining 3 models. 

\subsection{Results of MNIST-FASHION}
\begin{figure}[htbp]
    \centering
    \includegraphics[width=\columnwidth]{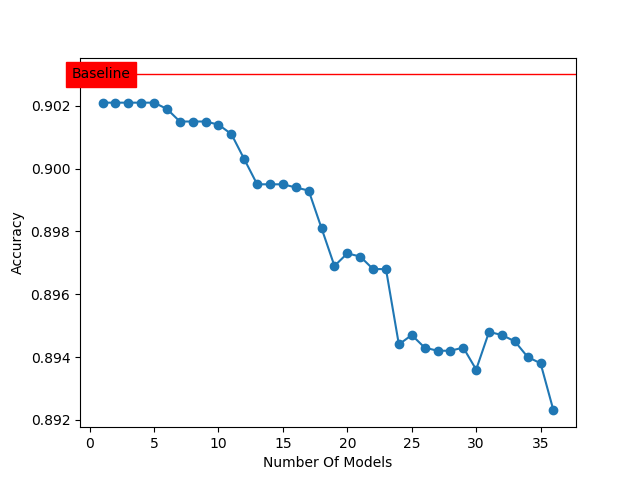}
    \caption{MNIST-FASHION}
    \label{mnist_results}
\end{figure}

Fig.~\ref{mnist_results} provides the results obtained on MNISH-FASHION testing set. Unlike the results from CIFAR-10 shown in Fig.~\ref{cifar10_results}, none of the composed ensembles were able to outperform the baseline model. These results are likely due to the fact that the baseline model is already performing very well (90.3\% accuracy on the testing set); thus none of the proposed combinations of models were able to achieve better performance. However, some of the ensembles with even a small number of models [1--5] have scored similar accuracy to the baseline model (90.21\%). This clearly shows that the real solution's $2^{36}-1$ is likely to have a set of ensembles that can produce promising results.

\subsection{Computational cost}
Further analysis for \emph{EnSyth} shows that the ensembles require less mathematical operations to make a prediction based on the trainable parameters count (assuming the number of parameters for an ensemble is equal to the model that has the maximum number parameters in that ensemble) and CPU execution time.
Table.~\ref{tab2} displays the number of trainable parameters and the average CPU execution time for the solution's space. CPU execution time was calculated as follows. Each model generated from a set of multiple solution spaces was fed with 50 randomly selected images from CIFAR10, and the same for MNIST-FASHION, and 25 images from CIFAR5. The experiment is repeated nine times, and then the average is calculated.

\begin{table}[htbp!]

\caption{Processing Time}
\begin{center}
\setlength\tabcolsep{4.2pt} 
\begin{tabular}{|c|c|c|c|c|c|c|}
\hline
\textbf{Solution}&\multicolumn{2}{|c|}{\textbf{CIFAR10}} &\multicolumn{2}{|c|}{\textbf{CIFAR5}}&\multicolumn{2}{|c|}{\textbf{MNIST-FASHION}} \\
\cline{2-7} 
\textbf{Space} & PARAMS & CPU & PARAMS & CPU & PARAMS & CPU \\
\hline
Baseline&1663367&91389&1663367&65342&1068298&60462\\
\hline

model1&449125&88712&495739&55458&690727&64838\\
model2&434727&97337&462392&57717&515694&63037\\
model3&409820&92000&413032&53179&349340&54986\\
model4&389077&104112&377410&63897&261155&63275\\
model5&372406&104244&350174&54509&216367&55309\\
model6&357104&95805&327820&58771&186753&57215\\
model7&300644&89662&262180&57328&124971&71198\\
model8&264190&87792&230782&59406&103398&68252\\
model9&238728&90467&210362&50445&89496&62281\\
model10&220266&98415&191688&63527&81341&68114\\
model11&202498&96473&173373&58674&73196&71749\\
model12&183802&87632&156328&58764&67678&60940\\

\hline
model13&449591&84103&495625&49144&690725&61207\\
model14&435068&87470&462176&61775&515985&71292\\
model15&409863&81002&412918&62938&349510&57813\\
model16&389414&87186&377683&53196&263364&66171\\
model17&372045&82882&350011&54519&216602&65960\\
model18&357415&89725&328173&57192&186321&63897\\
model19&125306&87472&262669&63158&300819&65514\\
model20&264433&91569&230848&59841&103440&65623\\
model21&238866&83084&210599&65157&89636&80378\\
model22&220109&85930&191402&62459&81384&72368\\
model23&203311&94199&173371&64266&73173&71940\\
model24&182823&82880&155159&47629&67589&69643\\

\hline
model25&449578&89749&495692&65576&690728&64918\\
model26&435326&84622&462402&57962&516312&70715\\
model27&410650&93787&413068&65927&349515&88424\\
model28&389634&92072&377807&56507&262603&69605\\
model29&372273&92882&350256&61985&216909&64090\\
model30&357602&87173&327886&57042&186680&78473\\
model31&300925&88574&262159&60170&125250&61816\\
model32&264239&91445&230685&58797&103346&67042\\
model33&239254&90179&210282&58426&91338&67711\\
model34&219892&89763&190855&52055&81142&68332\\
model35&202639&95498&173020&63604&73265&72857\\
model36&184180&94243&157021&56694&67622&67416\\

\hline
\end{tabular}
\label{tab2}
\end{center}
\end{table}

As shown in Table.~\ref{tab2}, all of the solution space models have a smaller number of trainable parameters than the baseline model in the three data sets. Besides, CPU execution time for most of the outperforming ensembles is less than the original model, where CPU execution time for an ensemble is the average CPU execution time in microseconds ($\mu$s) for the models that made up that ensemble. As a result, in a parallel execution environment, the generated ensembles will be faster in classifying an image and outperform the baseline model in terms of inference time and accuracy.
However, the number of parameters and the average CPU execution time are indirect performance indicators for \emph{EnSyth}, deeper investigation in a parallel execution environment is required in the future.

\subsection{Discussion} 
Besides our aim to generate better-compressed classifiers, we are particularly interested in exploring the space of the existing solutions in depth to identify the optimal combination of candidates that could be synthesised to produce better results. With only 36 models and a simple selection technique like backward elimination, \emph{EnSyth} was able to search a small subset of the space (only 36 out of $2^{36}-1$ possible solutions) to find one or more ensembles that outperforms the baseline model. 
As such, there is a very promising potential for further investigation including: (1) expanding the space of the pruned models by applying different pruning methods with different hyperparameters; and (2) dealing with the selection of the optimal ensemble as a multi-objective optimisation problem to find the minimum number of models (a small ensemble) while achieving the highest possible accuracy. Furthermore, the results from CIFAR10, CIFAR-5, and MNIST-FASHION indicate that with a small number of models in an ensemble, the classifier can achieve high predictability levels. This is particularly interesting when it comes to deploying those ensembles on smartphones and Internet of Things (IoT) devices, because the ensemble size is small compared to the baseline model and the performance in terms of inference time and accuracy is even better.

\section{Conclusion and future work} 
\label{conclusion}
In this paper, we describe \emph{EnSyth}, a method to synthesise deep learning ensembles, resulting in improved classifiers in terms of inference time and accuracy; first, we train a baseline model then prune it using different values for hyperparameters to form a pool of compressed models. After that, we employ ensemble learning to compose new predictors and search the solution space for outperforming models. The simulation results on CIFAR-10, CIFAR-5 datasets reflect the effectiveness of the proposed method to enhance the performance of compressed models.
At present, we depend on majority voting to create ensemble predictions with a relatively small number of models. In future, we will rely on more complex ensemble learning techniques such as stacking and boosting, also try to test the potential of this approach on a broad set of deep learning models and datasets to validate it on a larger space of solutions. Moreover, multi-objective optimisation using evolutionary algorithms can be considered to find the smallest ensemble with the best accuracy in the solution space. In addition, we aim to deploy the outperforming classifiers on real-time latency applications like a self-driving car where multiple models could be used for direct observation of driving. Another interesting application could employ a set of those models on resource-limited devices to collaboratively learn a shared prediction model that allows better accuracy, real-time inference and less power consumption levels.

\bibliographystyle{IEEEtran}
\bibliography{ensyth1}

\end{document}